\documentclass[letterpaper, 10 pt, conference]{ieeeconf}  

\IEEEoverridecommandlockouts                              

\usepackage{epsfig} 
\usepackage{amsmath} 
\usepackage{amssymb}  
\usepackage[caption=false,font=footnotesize,captionskip=1pt,farskip=3pt]{subfig}
\usepackage{todonotes}
\usepackage[T1]{fontenc} 
\usepackage{siunitx}
\usepackage{graphicx}
\usepackage{hyperref}

\let\ACMmaketitle=\maketitle
\renewcommand{\maketitle}{\begingroup\let\footnote=\thanks \ACMmaketitle\endgroup}

\makeatletter
\newcommand*\titleheader[1]{\begingroup\gdef\@titleheader{#1}\let\footnote=\thanks\endgroup}
\AtBeginDocument{%
  \let\st@red@title\@title
  \def\@title{%
  \begin{flushleft}
    \vspace{-2.0em}
    \bgroup\normalfont\small\@titleheader\par\egroup
    \vspace{-18pt}\par\noindent\rule{\textwidth}{0.1pt}
    \end{flushleft}
    \vskip0.5em\st@red@title
        }

}

\title{\LARGE \bf
Learning to Play Foosball: System and Baselines
}

\author{Janosch Moos$^{1}$, Cedric Derstroff$^{2}$, Niklas Schröder$^{3}$ and Debora Clever$^{4}$
\thanks{$^{1}$Janosch Moos is with the Institute for Mechatronic Systems, TU Darmstadt,
        64287 Darmstadt, Germany
        {\tt\small janosch.moos@tu-darmstadt.de}}%
\thanks{$^{2}$Cedric Derstroff is with the Hessian Center for Artificial Intelligence (hessian.AI) 64293 Darmstadt, Germany, and the Department of Computer Science, TU Darmstadt, 64283 Darmstadt, Germany
        {\tt\small cedric.derstroff@tu-darmstadt.de}}%
\thanks{$^{3}$Niklas Schröder is a student in the Department of Electrical Engineering,
        TU Darmstadt, 64283 Darmstadt, Germany
        {\tt\small niklas.schroeder@stud.tu-darmstadt.de}}%
\thanks{$^{4}$Debora Clever is with the Institute for Mechatronic Systems, TU Darmstadt,
        64287 Darmstadt, Germany and ABB AG, 68309 Mannheim, Germany
        {\tt\small debora.clever@ims.tu-darmstadt.de}}%
\thanks{\newline
© 2024 IEEE.  Personal use of this material is permitted. Permission from IEEE must be obtained for all other uses, in any current or future media, including reprinting/republishing this material for advertising or promotional purposes, creating new collective works, for resale or redistribution to servers or lists, or reuse of any copyrighted component of this work in other works.}
}

\titleheader{\textit{Accepted as a conference paper at the 2024 IEEE International Conference on Robotics and Automation (ICRA 2024)}}

\begin{document}

\maketitle
\thispagestyle{empty}
\pagestyle{empty}

\begin{abstract}
%
This work stages Foosball as a versatile platform for advancing scientific research, particularly in the realm of robot learning. We present an automated Foosball table along with its corresponding simulated counterpart, showcasing a diverse range of challenges through example tasks within the Foosball environment. Initial findings are shared using a simple baseline approach.
Foosball constitutes a versatile learning environment with the potential to yield cutting-edge research in various fields of artificial intelligence and machine learning, notably robust learning, while also extending its applicability to industrial robotics and automation setups.
To transform our physical Foosball table into a research-friendly system, we augmented it with a 2 degrees of freedom kinematic chain to control the goalkeeper rod as an initial setup with the intention to  be extended to the full game as soon as possible. Our experiments reveal that a realistic simulation is essential for mastering complex robotic tasks, yet translating these accomplishments to the real system remains challenging, often accompanied by a performance decline.
This emphasizes the critical importance of research in this direction. In this concern, we spotlight the automated Foosball table as an invaluable tool, possessing numerous desirable attributes, to serve as a demanding learning environment for advancing robotics and automation research.
\end{abstract}

\section{INTRODUCTION} \label{sec:intro}
Despite the progress made in robot learning, solving meaningful robotic tasks with artificial intelligence in real world scenarios is still a difficult problem. Large amounts of publications only discuss results of experiments on simulated environments. However, while the results may look promising, the transfer to real systems is another leap to take. A major reason are uncertainties of varying sources, such as the sim-to-real gap or state estimation. Especially when using vision based sensing, estimated poses of objects within an environment can be inaccurate for varying reasons, e.g., blurring, motion or insufficient camera resolution. However, uncertainties can be tackled by robust algorithms, which are specifically designed to minimize the impact that uncertainties have on the decision making policy. By design, robust algorithms have close ties to adversarial reinforcement learning, where two players learn opposite goals. In this case, the adversary is a source of uncertainty within the environment. With this work, we aim to build a robot learning environment, both in reality and simulation, as a research platform for robust and adversarial robot learning research. We further aim to provide preliminary learning results which are validated on the real system to serve as a baseline for said research. Therefore, the goal is an environment that is highly dynamic and allows for human robot interaction while applying a reasonable amount of restrictions to keep learning feasible. The environment should also have similar uncertainties to complex robotic environments. An application that meets these criteria is autonomous Foosball.

\begin{figure}
    \centering
    \includegraphics[width=\linewidth]{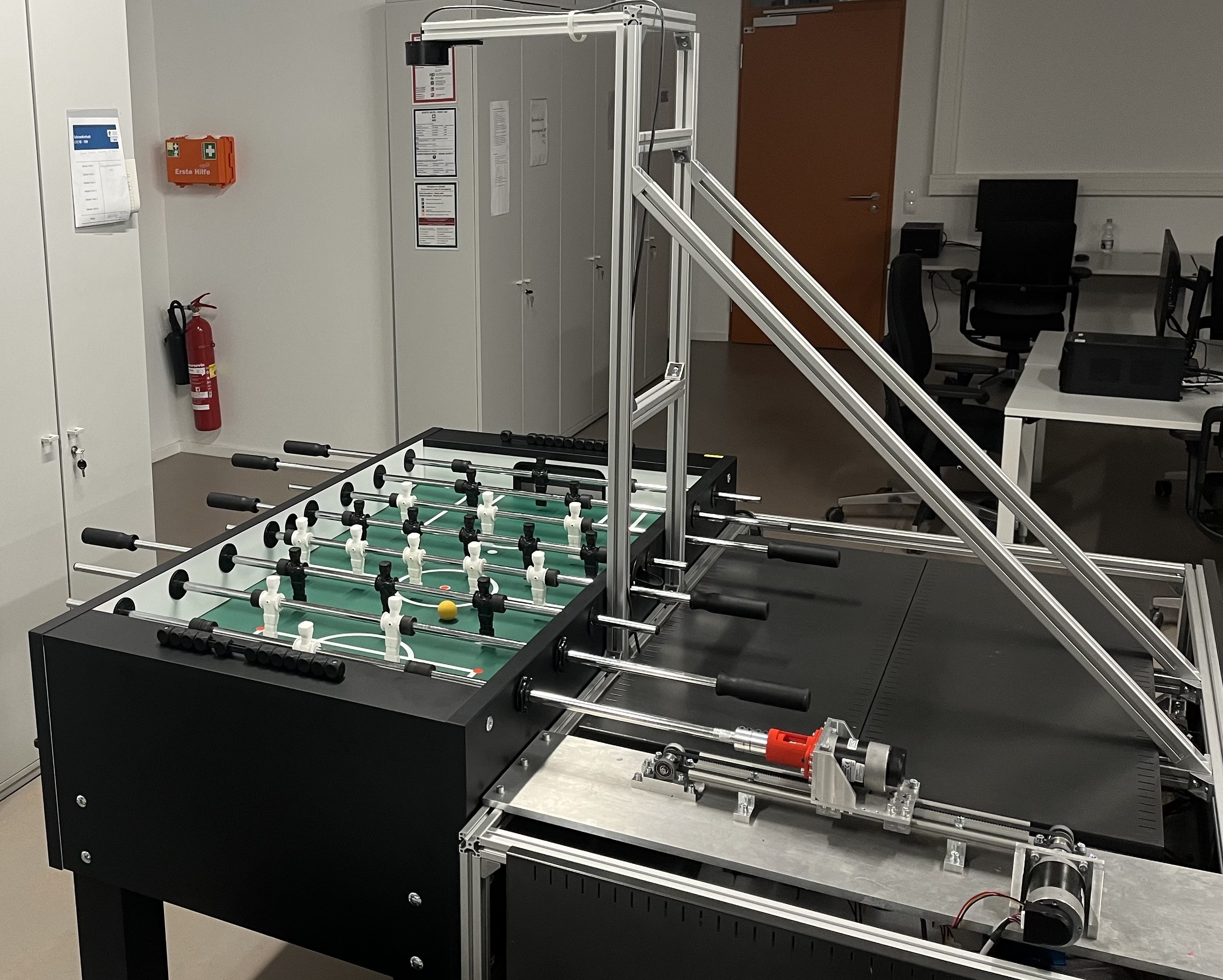}
    \caption{Our autonomous Foosball system in a human versus machine setting for robust and adversarial robot learning research.}
    \label{fig:real_table}
\end{figure}

Foosball, also known as table soccer, is a two (1\,vs.\,1) or four (2\,vs.\,2) player zero-sum game that emulates a soccer game. In our setup, we consider the most widely known version of a Foosball table which consists of a miniature soccer field with four different controllable rods for each team. As shown in Fig.~\ref{fig:real_table}, a total of 11 figurines, resembling soccer players, are spread across the four rods of each team. These figurines represent goal keeper (1 figurine), defense (2 figurines), midfield (5 figurines) and offense (3 figurines). Each rod has two \textit{degrees of freedom} (DoF) for translatory and rotary movements with mechanical joint limits in the prismatic and rule-based limits in the revolute joint. In autonomous Foosball, the rods of one team are automated to create a human vs. machine environment. By removing rods from each team, this environment is scalable between a cumulative \numrange{2}{8} DoF. During the game, the ball can reach velocities up to \qty{15}{m/s} which makes Foosball a highly dynamic and fast paced game and a challenging problem for robot learning. However, the game also imposes natural restrictions: objects of interest, i.e., ball or figurines, are limited to specific areas and movements. Further, interactions with humans happen within predefined bounds through the movement of opponent rods and the occasional retrieval of the ball. As such, autonomous Foosball is an excellent stepping stone towards complex robot learning applications as well as robust and adversarial learning research. However, while for our research, the properties concerning robust learning are most interesting, the Foosball table opens a lot more opportunities. Implied by the previous paragraph, Foosball constitutes a challenging and scalable multi-agent learning environment which includes, but is not limited to, team games with a human partner also translating to human robot interaction and intention modelling. In addition, Foosball provides a learning environment for intuitive and compact robotic skill representations, imitation learning, as well as real-time planning in continuous state and action spaces. Lastly, it is also worth mentioning that solving the Foosball environment strongly depends on auxiliary tasks such as high frequency state estimation and object tracking.

\section{BASICS}
\subsection{Adversarial RL}
In standard, single-agent Reinforcement learning (RL), the problem setup is usually formulated as a Markov decision process (MDP). Therefore, the agent operates in an environment which represents the task at hand; It observes the state $S_t\in \mathcal{S}$ at every discrete time step $t$ and takes an action $A_t \in \mathcal{A}$ based on its decision making rule, the policy, $\pi(S_t)$. With the application of this action, the environment transitions to the next state $S_{t+1}$ and the agent is rewarded by $R_t$. In episodic tasks like Foosball, this cycle continues until a terminal state is reached. The agent's objective is to maximize the (discounted) cumulative reward \cite{sutton2018}.

In multi-agent RL (MARL), the problem setup is usually modelled as a more complex Markov game, where the next state and rewards do not only depend on the actions of a single agent but on the \textit{joint} action.
Since the rewards of the agents might not be aligned, their objectives might neither.
In order to avoid this tremendous increase in complexity, one can also treat all other agents as part of the environment and consider different agent setups as domain randomization. In the MARL community, this is referred to as independent learners \cite{Bowling2000}.

This concept is especially applicable in adversarial settings, where the other agents are opponents, i.e., the rewards are at least partially opposite.
A common special case of adversarial settings are two-player zero-sum games which have been widely studied in the RL community \cite{Uther1997} but even much earlier in the area of game theory \cite{Owen1982}. Two-player zero-sum games are a common type of problem found in games, where "games" refers to the definition found in economics and extends far beyond just board and video games. As the name suggests, in two-player zero-sum games, there are two agents acting in an environment with opposite goals. Any gain an agent acquires results in an equivalent loss for the other, such that the sum of rewards for both agents is always zero \cite{Owen1982}. In the adversarial setting, the agents are often referred to as \textit{protagonist} and \textit{opponents}, where the protagonist is the agent we are training.

When training an agent in an adversarial setting, e.g., two-player zero-sum games, we need to take into account that the opponent maximizes its expected reward which in turn minimizes the protagonist's score. To train efficiently in such a setting, the protagonist needs an opponent of appropriate skill level. Consequently, when its own skill level increases, the opponent must also become stronger in order to be challenged and not to overfit that performance.

To apply this mechanism to reinforcement learning, we can make use of the self-play paradigm. Self-play means that the protagonist learns the task by playing against copies of itself. Thus, the opponent is either an exact copy of the current version of the protagonist or an older version of itself. Replacing the opponent by an updated version of the protagonist every once in a while yields a new opponent of adequate strength and guarantees a challenging learning environment. This methodology can also be seen as a specialized version of auto-curriculum learning.

Self-play has been widely used for a long time and has been successfully applied to a range of challenging domains \cite{silver2017,tesauro1995,samuel1959,baker2020,openai2019,alshedivat2018,bansal2018}. While self-play is a na\"{i}ve approach to adversarial learning with the protagonist being an independent learner, it still serves as a baseline for more complex approaches down the line.

\subsection{Related Work} \label{sec:related}
Foosball has been considered in scientific contexts before. In \cite{Janssen2010} and \cite{Janssen2012}, the authors discuss tracking of the game ball through visual cues. Senden et al. apply invariant world models for robust object localization in environments where a camera sensor is not static \cite{Senden2022}. \textit{KiRo} \cite{Weigel2003}, commercially known as \textit{Starkick}, is a complete autonomous Foosball system and one, if not the first, of its kind. Other examples exist mainly in the context of university projects. However, none of these examples employ AI-based controllers but rely on a set of predefined actions. To the best of our knowledge, there are only few papers that discuss AI-based approaches in the context of Foosball. The first is KIcker \cite{De2021}, a demonstrator for industrial application of deep reinforcement learning utilizing a digital twin to a real system and domain randomization. The learning environment, however, is restricted to the basic example of shooting goals with the offense. A second work \cite{Rohrer2021} applies Deep Q-Learning to block incoming shots with the goal keeper both in simulation and on a real system. However, in their work the authors only consider a set of \num{3} actions which poses a significant restriction. Another work \cite{Gashi2023} has been published only two weeks prior to this submission. The authors discuss the use of multi-agent competition to train AIs for offense and keeper and theirs is the only other work considering the opponent in its strategy. The presented approach is the most sophisticated of the prior work and is a direct extension of \cite{De2021} and \cite{Rohrer2021}. However, their work does not consider a continuous game of Foosball. In contrast, our work aims for a more complete picture with a direct application to future research in robust and adversarial learning.

\section{Learning to Play Foosball}
Foosball as an environment varies depending on the table used. Most tables differ in size, shape of the figurines, color, etc. In our work, we build upon an \textit{Ullrich ``Home''} table which follows the configuration described in Sec.~\ref{sec:intro} with a playing field of \qtyproduct{1.2 x 0.68}{\m}.

\subsection{Our Autonomous Foosball System} \label{sec:our_system}

\begin{figure}[b]
    \centering
    \includegraphics[width=\linewidth, trim=0 24 0 10, clip]{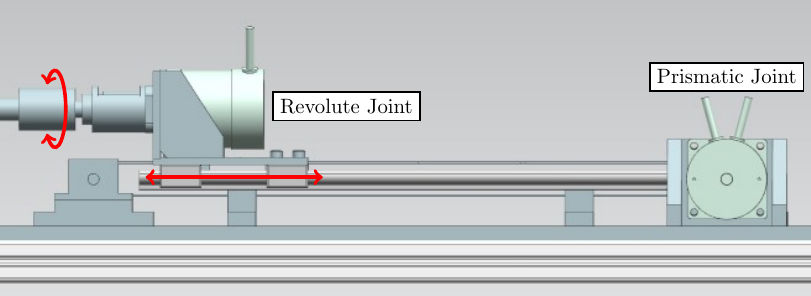}
    \caption{Shown is the CAD model of the actuation for the autonomous Foosball system. A BLDC motor operates a belt drive system with a carriage creating a prismatic joint. A second BLDC motor is mounted onto the carriage and connected to the rod to add a revolute joint.}
    \label{fig:CAD_Actuation}
\end{figure}

With autonomous Foosball, we aim to automate a subset of the table's rods to create a human vs. machine setting. Each automated rod needs actuation in 2 DoF, i.e., translation and rotation. In our system, a \textit{brushless DC} (BLDC) motor is connected to a belt drive with a carriage moving on a guiding rail to create a prismatic joint. For reference, a CAD sketch is included in Fig.~\ref{fig:CAD_Actuation}. Mounted on the carriage is a second BLDC motor as a revolute joint with a direct link to the rod of the Foosball table creating a 2 DoF open kinematic chain. To protect the motor axis of the revolute joint during high accelerations of the prismatic joint, a thrust bearing redirects forces imposed by the rods away from the motor axis and onto the carriage instead. Given the specifications of the actuated rod, belt drive and carriage, as well as, measuring trajectories of a human player, a rough estimate for the motor requirements of both movements is calculated. The estimate is further adjusted to fit desired criteria, i.e., achievable ball speeds through consideration of energy conservation during a shooting motion. While the highest recorded ball velocity is over \qty{15}{m/s}, a regular game rarely exceeds \qty{10}{m/s}. Therefore, the hardware is designed to accommodate speeds of up to \qty{10}{m/s}. For our application we utilize \textit{Nanotec's DB59L048035} motor for the revolute and \textit{DB59C048035} for the prismatic joint in combination with their \textit{C5-E} motor controllers. Currently, the automation is restricted to the white goal keeper for initial testing and validation purposes and will be extended to all four rods of one team in the near future.

\begin{figure}[b]
    \centering
    \includegraphics[width=\linewidth,page=5, trim=0 10 0 5, clip]{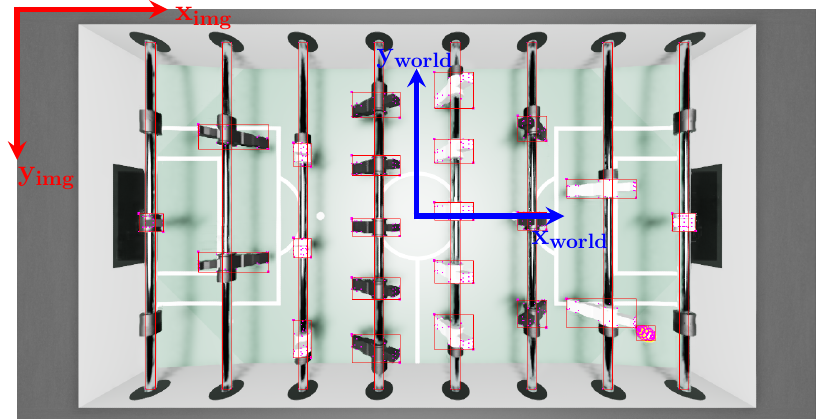}
    \caption{Shown is an example of synthetically generated data used to train the detection algorithm YOLO used in this work. The bounding boxes are created from a set of points projected into the image plane. These points are taken from the underlying CAD model of the figurines.}
    \label{fig:Synthetic_Data}
\end{figure}

While encoder based readings in the motors provide a reliable position and velocity estimation of the automated rods, obtaining the same for the ball and opponent rods is more difficult. In case of the opponent rods, the minimum required information is the translatory position to avoid being blocked during attacks. In addition, the position and velocity of the ball must be estimated. In our system, a vision-based state estimation is employed. A camera is mounted above the table, as shown in Fig.~\ref{fig:real_table}, with a resolution of \numproduct{1280 x 720} pixels at up to \num{90} frames per second (FPS). Prior work has applied traditional detection schemes through, e.g, Hough transformations and color coding to detect the ball while opponent figurines are mostly ignored  \cite{Janssen2010, Janssen2012, Senden2022}. To design a state estimator that provides more information compared to most prior work while also being applicable to more complex robotic environments later on, we use the state-of-the-art detection algorithm \textit{You only look once} (YOLOv5s) \cite{Redmon2016, Jocher2020} as it provides flexible high frequency object detection. However, a drawback of such a learning based detection algorithm is the data requirements and generating labeled data from real world images is a tedious, time-consuming process. In our work, we utilize the simulated learning environment used to train intelligent robot controllers to also create synthetic data to train YOLO. The simulated learning environment is generated in NVIDIA's Isaac Sim \cite{Makoviychuk2021}, a GPU-based physics simulator with realistic graphics. The simulation recreates the camera system virtually to provide more realistic synthetic training data. An example of the synthetic data is shown in Fig.~\ref{fig:Synthetic_Data}.

\begin{figure*}[t]
     \centering
    \subfloat[Keeper Position Tracking]{%
         \label{fig:fig_pos_est}%
         \includegraphics[width=0.33\textwidth,page=2]{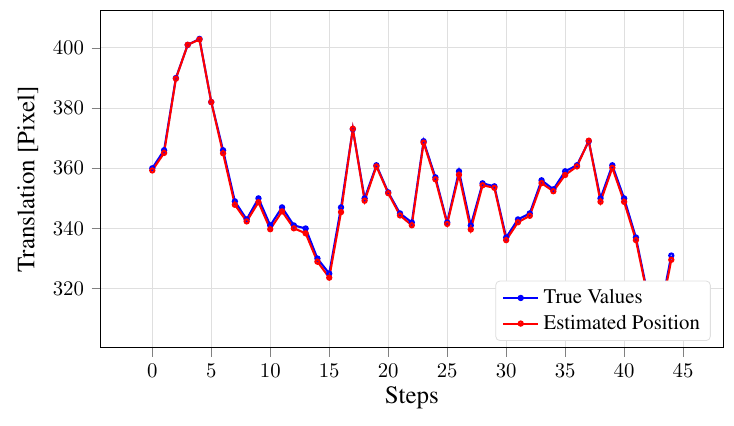}%
    }%
    \hfill%
     \subfloat[Ball Position Tracking]{%
         \label{fig:ball_pos_est}%
         \includegraphics[width=0.33\textwidth,page=3]{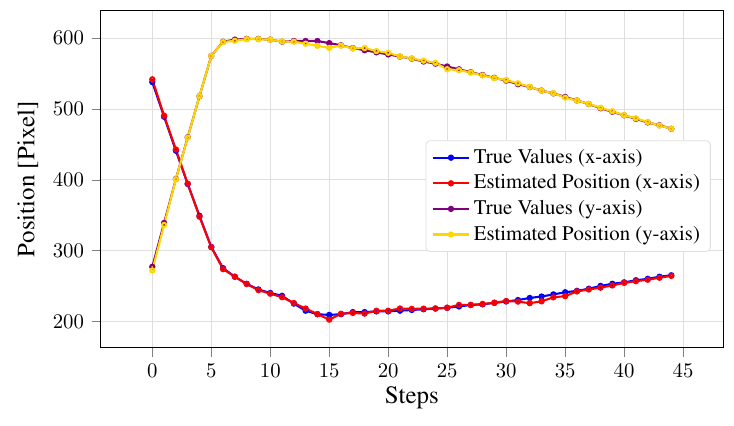}%
     }%
    \hfill%
     \subfloat[Ball Velocity Estimation (x-Axis)]{%
          \label{fig:ball_vel_est}%
         \includegraphics[width=0.33\textwidth,page=4]{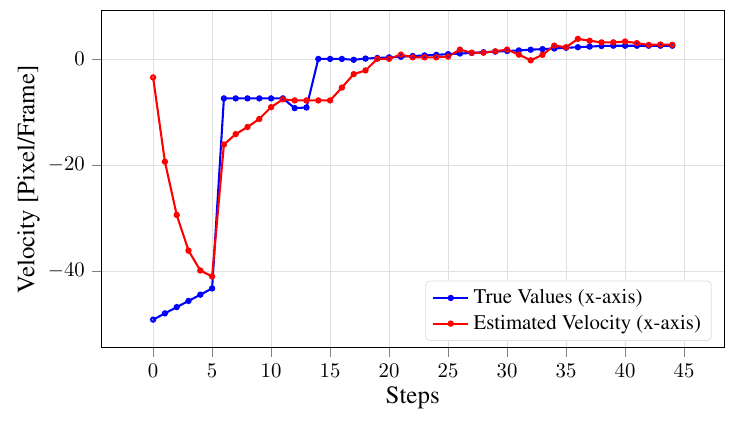}%
    }
    \caption{Shown is a subset of the state estimation results from YOLO detection followed by Kalman filtering in comparison the the true values from simulation. All estimates are in pixels or pixels per frame at \num{60} FPS. The white keeper rod is chosen in (a) as representative for position estimation along its prismatic joint. Other rods display similar performance. In the center (b), the x- and y-position estimate of the ball is plotted. Lastly in (c), the ball velocity in x-direction is shown. We have omitted the y-velocity for clarity as it yielded similar results.}
    \label{fig:kalman_result}
\end{figure*}

Still, an object detector can only provide measures for the positions of the objects in the scene. Through combination with a Kalman Filter \cite{Kalman1960}, further information about velocities of the objects is retrieved. The image processing and decision making is then done on a remote computer and the computed motor commands are communicated to the low-level motor controllers via CAN-Bus. The Kalman Filter is further exploited to predict future states to reduce the impact of delays caused by computation and communication. The state estimation is tested in simulation to get a preliminary validation of the performance. The results are shown in Fig.~\ref{fig:kalman_result} with estimated positions and velocities in the scene in comparison to the true values. Due to the highly dynamic nature of Foosball, we aim to extend the Kalman Filter predictions to consider collisions of the ball with objects. So far, we have included collisions with the surrounding walls. Collisions with the figurines are still a work in progress as they depend on estimates of the figure rotations and rotational velocities. While early attempts have been made to estimate the rotations, the results are still preliminary and therefore excluded from this paper.

\subsection{Training Setup}
As mentioned in Sec.~\ref{sec:related}, most prior autonomous Foosball agents heavily rely on predefined deterministic actions. Only in few works \cite{De2021, Gashi2023} reinforcement learning is being discussed. In these works, learning is either limited to the most basic tasks or only done in simulation. We, on the other hand, want to train a strong agent from scratch that is capable of playing in an adversarial setting, both in simulation and reality. A first step towards this goal is a baseline approach relying on RL and self-play. As representative, we use Proximal Policy Optimization (PPO) \cite{schulman2017} as it is still one of the most widely used state-of-the-art RL algorithms.

\subsubsection{The Training Environment}
For our training, we utilize Omniverse Isaac Gym \cite{Makoviychuk2021} as the learning environment \footnote{\url{https://github.com/Jaykixx/Foosball.git}} where we have recreated our Foosball system through CAD modeling as shown in Fig.~\ref{fig:Synthetic_Data}. In the simulation, all rods, including the opponent rods, are each driven by a 2 DoF kinematic chain with a prismatic and a revolute joint. These joints are parameterized according to the motors used in our real system. The range of the prismatic joints matches the range of motion of the corresponding rod. The revolute joints are limited to two revolutions based on the rules of Foosball. Each joint can be controlled through position, velocity or torque targets. For this work, however, we use positions. All training is performed in task-specific settings from basic tasks, such as blocking incoming shots without opponent, to more complex self-play settings with either two (keeper\,vs.\,keeper) or all eight (full game) rods in place. This approach allows for varying levels of difficulty.

\subsubsection{The Observation Space}
Depending on the number and type of rods involved in the training process the observation space changes. For each active joint in a task, we observe joint position and velocity. In tasks involving opponents that we transfer to the real system, the observations of the opponents are limited to prismatic joints, as explained in Sec.~\ref{sec:our_system}. Optionally, instead of the joints, it is also possible to observe the figurines, which may be beneficial for some tasks. In addition to the joints or figurines, the agents observe the position and velocity of the ball in the x- and y-directions in world coordinates as defined by Fig.~\ref{fig:Synthetic_Data}. 

\subsubsection{The Rewards} \label{sec:rewards}
Rewarding and punishing the agent only for shooting or catching goals, respectively, gives the agent the most freedom to learn meaningful skills and strategies. The downside is, however, a very sparse reward problem which makes it extremely difficult to learn beneficial behavior. Thus, we studied different ways to speed-up training by also rewarding agents in other situations through reward shaping. A consequence is less freedom and added bias in the solution. The base reward is a positive or negative integer for win or loss. In addition, a penalty is given when the ball flies upwards out of the table. We further use three different auxiliary rewards for reward shaping that are added as necessary for the different tasks. We reward closing the distance between ball and opponent goal, which is beneficial in tasks where most rods are active and goals become more rare. The second adds rewards for minimizing the distance between the closest figurine on a rod to the ball in the direction of the prismatic joint. This reward helps to increase the amount of interaction with the ball during early stages of the training process. Lastly, we have included action regularization to reduce unnecessary motion and thus load on the hardware during real system testing.

\section{EXPERIMENTS AND RESULTS}
As described in the previous sections, we have trained agents on a variety of different tasks that are relevant to Foosball, including both base skills and self-play. We use a neural network policy with 3 hidden layers with $[256, 128, 64]$ neurons. Policy outputs are limited to a range of $[-1, 1]$ which are then scaled up to match the joint limits for position control. Training runs on \num{8192} parallel environments. The PPO updates consider a horizon of \num{16} for base skills and \num{64} for self-play. Each update consists of \num{8} mini epochs with mini batches of either \num{16384} or \num{65536} depending on the horizon. The PPO objective is clipped to \num{0.2} with $\gamma = 0.99$ and $\lambda = 0.95$ for the generalized advantage estimate. As learning rate we are using an initial value of \num{1e-3} with either an adaptive scheduling or none at all. Both training and simulation run on a single NVIDIA GeForce RTX 3090 GPU with \qty{240}{Hz} for the physics and \qty{60}{Hz} for the decision making. In the experiments, we always observe the joint positions, not the figurines. For the discussion of the results, we have split the tasks between base skills and self-play. Goals give a reward of \num{\pm1000} depending on which goal was hit in all tasks.

\subsection{Base Skills}
As a first starting point, the simulation was used to learn a set of base skills that are part of the Foosball game, e.g., blocking and scoring.

\begin{figure*}[t]
    \centering
    \subfloat[Successful completions of the objective]{%
         \label{fig:base_success}%
         \includegraphics[height=4.5cm]{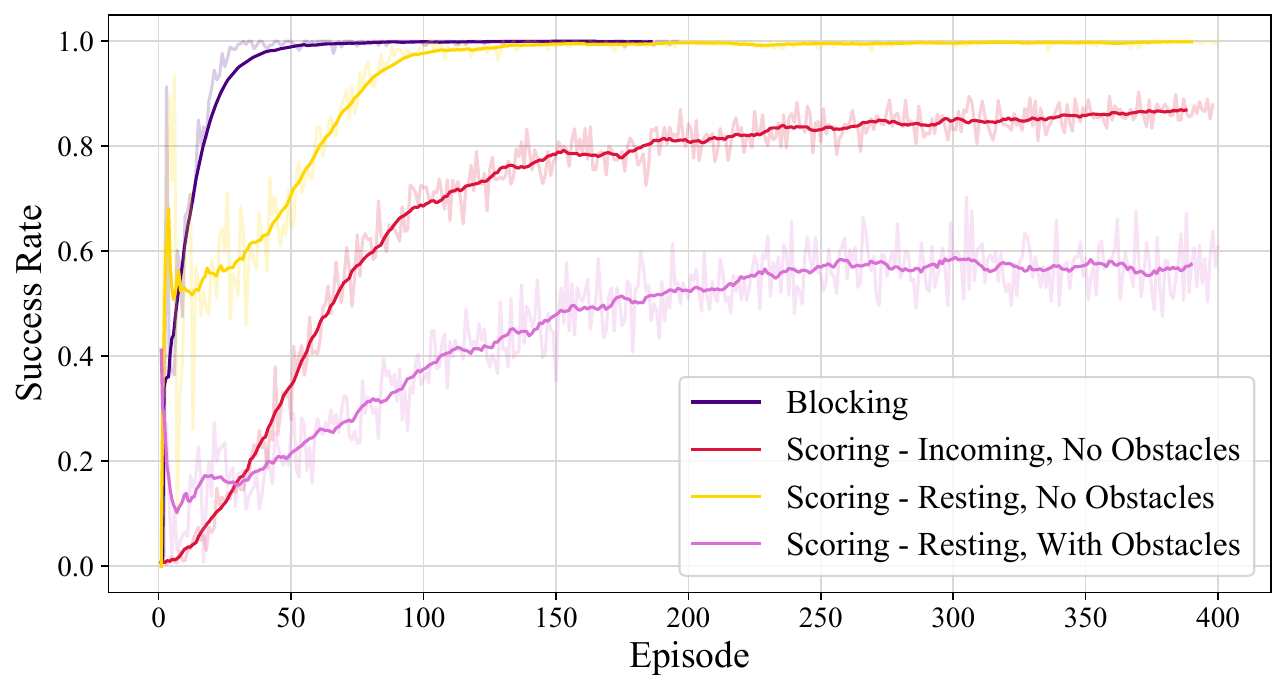}%
    }%
    \hfill%
     \subfloat[Rewards during base skill training]{%
         \label{fig:base_rewards}%
        \includegraphics[height=4.5cm]{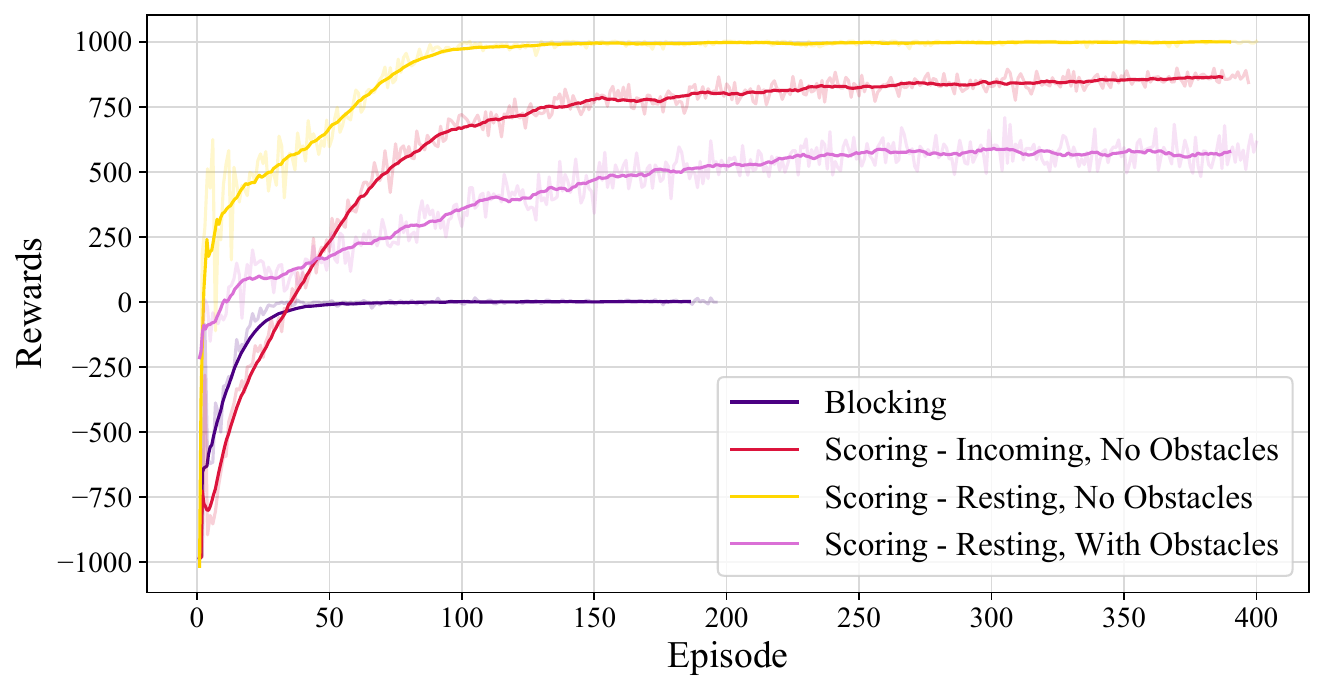}%
    }%
    \\
    \subfloat[Win and loss rates of the self-play agent]{%
         \label{fig:sp_win}%
         \includegraphics[height=4.5cm]{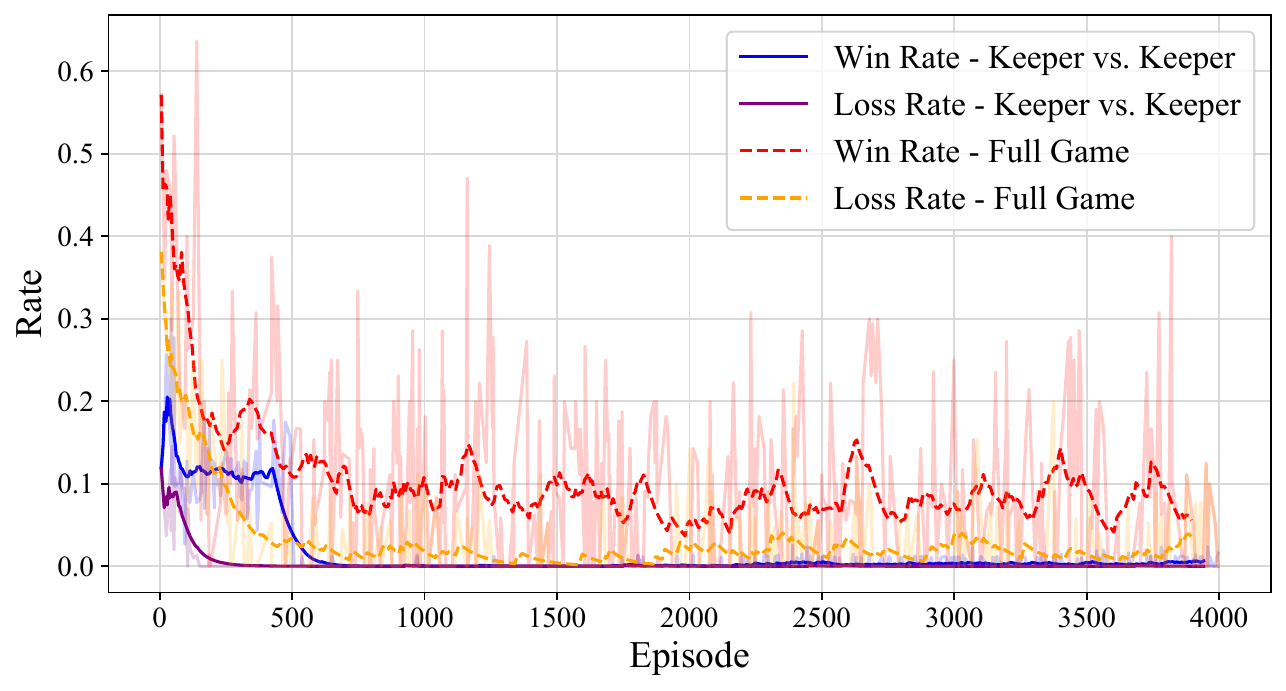}%
    }%
    \hfill%
     \subfloat[Rewards during self-play training]{%
         \label{fig:sp_rewards}%
        \includegraphics[height=4.5cm]{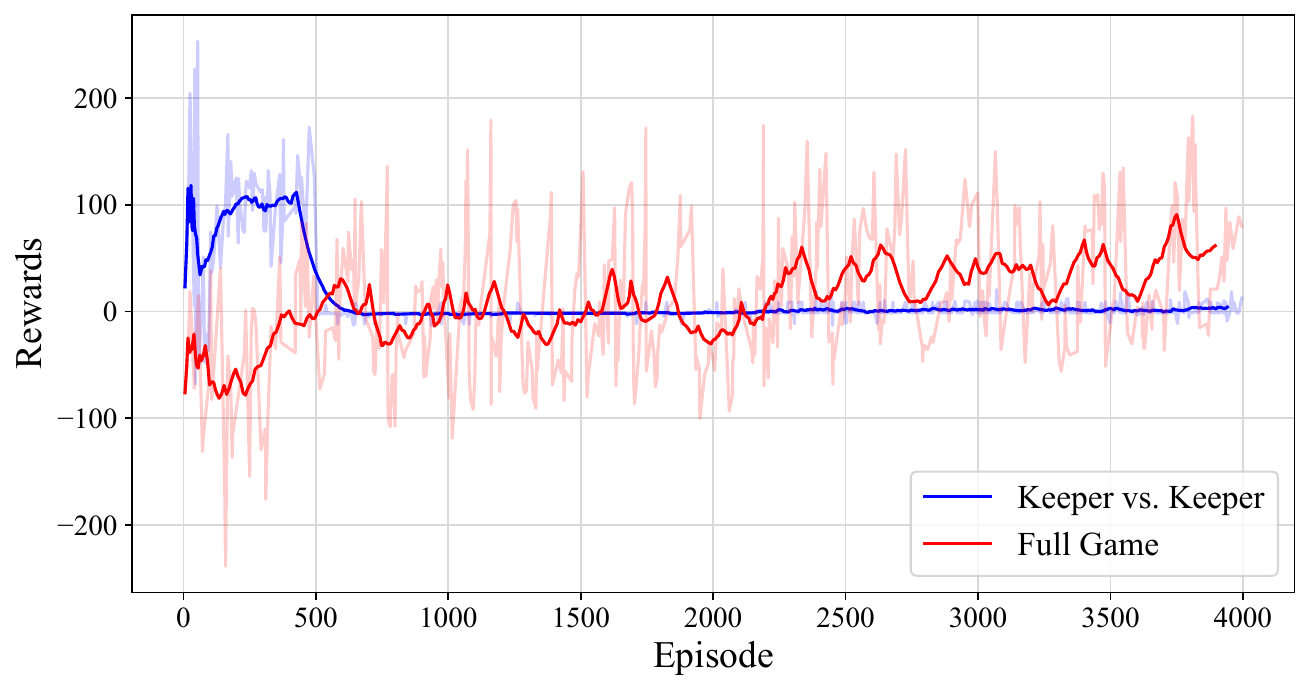}%
    }%
    \caption{Shown are the learning curves on our example tasks. The top row consists of the results on the base skills, while the bottom row illustrates the results of the self-play training. In each case, the left subplot is in terms of success rate, and the right subplots shows the reward over learning episodes.}
    \label{fig:base_skills}
\end{figure*}

\subsubsection{Blocking}
The blocking task is designed to teach the AI to block incoming balls with the goal keeper. The ball is set to a random starting location on the opponents side of the playing field with a random velocity of \SIrange{2}{7}{m/s} towards a random position within the goal. While the hardware is designed for velocities up to \qty{10}{m/s}, the velocity range in this task is more representative for the average player. The observations include ball position and velocity in x- and y-direction. Optionally, the prismatic position and or velocity of the keeper can be included for a total of \numrange{4}{6} dimensions. For the presented results, we have included the prismatic position. The output is a single action value for the prismatic joint. Fig.~\ref{fig:base_skills} shows the results of the training. As can be seen, the trained policy manages to block all incoming balls after less than \num{100} policy updates. However, the final policy has additionally been tested on our real system. There, we have noted that the policy had difficulties to react to fast shots that were consistently blocked in simulation. We have included video evidence in the supplementary material.

\subsubsection{Scoring}
Scoring as a task can be seen as an alteration of the blocking task through extension of the action space to the revolute joint. Given the rewards as described in Sec.~\ref{sec:rewards}, the AI will learn to repel the incoming balls into the opponent goal. Further variation can be added by initializing the ball in a resting position within a reachable distance of the motion limits of the keeper. In both cases, the observations extend to the position and or velocity of the revolute joint for a total of \numrange{4}{8} dimensions. For the presented results, we ignored the joint velocities. Policies are trained for either resting or incoming balls. As shown in Fig.~\ref{fig:base_skills}, scoring from resting ball positions without obstacles is easily mastered by the policy. In contrast, incoming balls are only repelled into the opponent goal about \qty{85}{\%} of the time even though all incoming balls are blocked.

Further difficulty is added by including opponent rods that remain static to serve as obstacles. In this case, the observations also include the prismatic positions of the added opponent rods which are randomized at the start of each run. For our experiments, we have added the keeper, defense and offense of the opponent as obstacles while always starting with a resting ball. Due to the increased difficulty, the success rate drops to around \qty{60}{\%}. Further analysis has shown that the trained policy ignores the possibility of redirecting the ball around obstacles by shooting against the walls and instead focuses on direct shots. However, due to the randomized positioning of both ball and obstacles, there is not always a direct path to the goal.

\subsection{Self-play}
The full potential of Foosball only unveils when introducing an opponent player. In other games, a variety of self-play techniques using different levels of planning and look ahead have been applied. The renowned AlphaZero \cite{silver2017} algorithm, for example, uses Monte-Carlo Tree Search to look ahead several steps and states in order to improve decision making over using the plain policy. In stochastic continuous problems like Foosball, this approach is not as easy. In this first step, we therefore solely rely on the policy for self-play playing against a past version of our agent, updating that opponent when winning against it reliably. We consider two variations of our environment for this approach, keeper\,vs.\,keeper and the full game.

\subsubsection{Keeper vs. Keeper}
At the time of submission, our real system features a single automated rod, i.e. the white goal keeper. Hence, we train the agent to compete in a fair setting where there are only the two goal keepers. The state estimation on the real system is also limited to observing the prismatic motion of the opponent rods. Therefore, we have equally limited the observations in the simulation. Compared to the base skills, however, the policy observes both its own position and velocity. As such, the observation space has a total of \num{10} dimensions. The ball starts in a randomized resting position within a reachable distance of either of the two keepers. The opponent is updated as soon as the protagonist wins at least \qty{20}{\%} of the games. In our experiments, the opponent is the same across all environments. However, optionally, a mix of multiple older version of the policy can be used as opponents. The environment is designed to be point symmetric such that the observations can be easily inverted for the opponent. The results of the training process are summarized in Fig.~\ref{fig:base_skills}\,\subref{fig:sp_win} and \subref{fig:sp_rewards}. We can see that the protagonist always stays slightly ahead in accordance with expectations. There is also a direct correlation between the updates of the opponent and the rewards/win rate of the protagonist. The physical distance between the two players causes the number of goals to drop significantly compared to the scoring task. 

\subsubsection{Learning the Full Game of Foosball}
Having shown that learning a simple adversarial version of the game of Foosball is feasible, this part will treat the training of the full game including all rods. The observations are in this case not limited such that all \num{16} joints and their velocities can be observed, increasing the observation space to \num{36} with \num{8}-dimensional actions per player. While the trained policy shows overall improvement in behavior, purposeful shooting patterns and other skills remain a sparsity. This result indicates that further training and improved algorithms and planning is required. 

\subsection{Transfer to the real system}
When applied on the real system, the performance of the policies visibly drops. One of the reasons is the policies overfitting to the observations seen in simulation. When the Kalman filter is added in simulation to estimate the ball state, instead of true values, the policies exhibit similar deterioration in performance. However, while training with Kalman filter estimates improves the quality of the policy on the real system, there is still a noticeable difference in performance compared to the simulation results. A human opponent could leverage these vulnerabilities to deliberately confuse the policy in game. These findings highlight the importance of robustness not just against uncertainties in the physical behavior of the system but also observations. We have included corresponding results of the discussed findings in the supplementary video. 

\section{CONCLUSIONS}
We have shown why a complex physics game like Foosball is worth studying and that it can be used as an intermediate step towards real world applications in robotics and automation as it features similar difficulties.
We have further presented how a Foosball table can be prepared for AI, robotics and automation research. In addition, we set up a lightweight realistic simulated version of said Foosball table using the NVIDIA Isaac Sim platform.
We highlighted the need for efficient realistic simulations to tackle robotics problems to minimize laborsome and time-consuming training on the real system.
In this concern, we analyzed the portability of several base skills as well as a keeper vs. keeper policy trained in simulation to the real Foosball table. Our findings demonstrate how vulnerable policies are to overfitting simulative environments both in observations and physical behavior, emphasizing the need for robust robot learning approaches. We further discuss the lack of complex strategies and purposeful behavior in na\"ive self-play. In a next step, the provided results may be improved by more sophisticated self-play, e.g., look-ahead planning with Monte Carlo tree search as well as robust learning schemes.
Further, we treated the opponent as part of the environment and therefore do not model intentions of any kind. This could, in an even further step, be combined by taking inspiration of professional Foosball players who try to infer intentions of the opponent by observing the opponents hand movements.
Speaking of visual input, the agent could also be trained end-to-end on raw pixels like commonly done in video games.
The full game of Foosball can also be extended to an adversarial team game where teams of two agents compete against each other as in real Foosball showcasing human robot collaboration.
To conclude, Foosball is a promising research platform for robotics and automation research and already shows promising results.
We hope our work will inspire blossoming future research contributing to many subareas of AI, ML and robotics.

\addtolength{\textheight}{-10.7cm}  





\section*{ACKNOWLEDGMENT}

We would like to thank \textit{ABB AG.} for their financial and specialist support.
We thank \textit{Ullrich Sport RiproTec GmbH} for providing the Foosball table as well as \textit{Nanotec Electronic GmbH \& Co. KG} for providing hardware components. This research project was partly funded by the Hessian Ministry of Science and the Arts (HMWK) within the projects ``The Third Wave of Artificial Intelligence - 3AI'' and hessian.AI.


\bibliographystyle{IEEEtran}
\bibliography{IEEEabrv,iros-2023}

\begin{thebibliography}{10}
\providecommand{\url}[1]{#1}
\csname url@rmstyle\endcsname
\providecommand{\newblock}{\relax}
\providecommand{\bibinfo}[2]{#2}
\providecommand\BIBentrySTDinterwordspacing{\spaceskip=0pt\relax}
\providecommand\BIBentryALTinterwordstretchfactor{4}
\providecommand\BIBentryALTinterwordspacing{\spaceskip=\fontdimen2\font plus
\BIBentryALTinterwordstretchfactor\fontdimen3\font minus
  \fontdimen4\font\relax}
\providecommand\BIBforeignlanguage[2]{{%
\expandafter\ifx\csname l@#1\endcsname\relax
\typeout{** WARNING: IEEEtran.bst: No hyphenation pattern has been}%
\typeout{** loaded for the language `#1'. Using the pattern for}%
\typeout{** the default language instead.}%
\else
\language=\csname l@#1\endcsname
\fi
#2}}

\bibitem{sutton2018}
R.~S. Sutton and A.~G. Barto, \emph{Reinforcement learning: An
  introduction}.\hskip 1em plus 0.5em minus 0.4em\relax MIT press, 2018.

\bibitem{Bowling2000}
M.~H. Bowling and M.~M. Veloso, ``An analysis of stochastic game theory for
  multiagent reinforcement learning,'' 2000.

\bibitem{Uther1997}
W.~Uther and M.~Veloso, ``{Adversarial Reinforcement Learning},'' 1997.

\bibitem{Owen1982}
G.~Owen, \emph{Game Theory}.\hskip 1em plus 0.5em minus 0.4em\relax Academic
  Press, 1982.

\bibitem{silver2017}
D.~Silver, T.~Hubert, J.~Schrittwieser, I.~Antonoglou, M.~Lai, A.~Guez,
  M.~Lanctot, L.~Sifre, D.~Kumaran, T.~Graepel, T.~Lillicrap, K.~Simonyan, and
  D.~Hassabis, ``Mastering chess and shogi by self-play with a general
  reinforcement learning algorithm,'' 2017.

\bibitem{tesauro1995}
\BIBentryALTinterwordspacing
G.~Tesauro, ``Temporal difference learning and td-gammon,'' \emph{Commun. ACM},
  vol.~38, no.~3, p. 58–68, mar 1995. [Online]. Available:
  \url{https://doi.org/10.1145/203330.203343}
\BIBentrySTDinterwordspacing

\bibitem{samuel1959}
A.~L. Samuel, ``Some studies in machine learning using the game of checkers,''
  \emph{IBM Journal of research and development}, vol.~3, no.~3, pp. 210--229,
  1959.

\bibitem{baker2020}
B.~Baker, I.~Kanitscheider, T.~Markov, Y.~Wu, G.~Powell, B.~McGrew, and
  I.~Mordatch, ``Emergent tool use from multi-agent autocurricula,'' 2020.

\bibitem{openai2019}
OpenAI, C.~Berner, G.~Brockman, B.~Chan, V.~Cheung, P.~Dębiak, C.~Dennison,
  D.~Farhi, Q.~Fischer, S.~Hashme, C.~Hesse, R.~Józefowicz, S.~Gray,
  C.~Olsson, J.~Pachocki, M.~Petrov, H.~P. d.~O.~Pinto, J.~Raiman, T.~Salimans,
  J.~Schlatter, J.~Schneider, S.~Sidor, I.~Sutskever, J.~Tang, F.~Wolski, and
  S.~Zhang, ``Dota 2 with large scale deep reinforcement learning,'' 2019.

\bibitem{alshedivat2018}
M.~Al-Shedivat, T.~Bansal, Y.~Burda, I.~Sutskever, I.~Mordatch, and P.~Abbeel,
  ``Continuous adaptation via meta-learning in nonstationary and competitive
  environments,'' 2018.

\bibitem{bansal2018}
T.~Bansal, J.~Pachocki, S.~Sidor, I.~Sutskever, and I.~Mordatch, ``Emergent
  complexity via multi-agent competition,'' 2018.

\bibitem{Janssen2010}
R.~Janssen, J.~de~Best, and R.~van~de Molengraft, ``Real-time ball tracking in
  a semi-automated foosball table,'' in \emph{RoboCup 2009: Robot Soccer World
  Cup XIII 13}.\hskip 1em plus 0.5em minus 0.4em\relax Springer, 2010, pp.
  128--139.

\bibitem{Janssen2012}
R.~Janssen, M.~Verrijt, J.~de~Best, and R.~van~de Molengraft, ``Ball
  localization and tracking in a highly dynamic table soccer environment,''
  \emph{Mechatronics}, vol.~22, no.~4, pp. 503--514, 2012.

\bibitem{Senden2022}
J.~Senden, K.~Jebbink, H.~Bruyninckx, and R.~Van De~Molengraft,
  ``Invariant-based world models for robust robotic systems demonstrated on an
  autonomous football table,'' \emph{IEEE Robotics and Automation Letters},
  vol.~7, no.~3, pp. 8542--8549, 2022.

\bibitem{Weigel2003}
T.~Weigel and B.~Nebel, ``Kiro--an autonomous table soccer player,'' in
  \emph{RoboCup 2002: Robot Soccer World Cup VI 6}.\hskip 1em plus 0.5em minus
  0.4em\relax Springer, 2003, pp. 384--392.

\bibitem{De2021}
S.~De~Blasi, S.~Kl{\"o}ser, A.~M{\"u}ller, R.~Reuben, F.~Sturm, and T.~Zerrer,
  ``Kicker: an industrial drive and control foosball system automated with deep
  reinforcement learning,'' \emph{Journal of Intelligent \& Robotic Systems},
  vol. 102, no.~1, p.~20, 2021.

\bibitem{Rohrer2021}
T.~Rohrer, L.~Samuel, A.~Gashi, G.~Grieser, and E.~Hergenr{\"o}ther, ``Foosball
  table goalkeeper automation using reinforcement learning.'' in \emph{LWDA},
  2021, pp. 173--182.

\bibitem{Gashi2023}
A.~Gashi, E.~Hergenr{\"o}ther, and G.~Grieser, ``Efficient training of foosball
  agents using multi-agent competition,'' in \emph{Science and Information
  Conference}.\hskip 1em plus 0.5em minus 0.4em\relax Springer, 2023, pp.
  472--492.

\bibitem{Redmon2016}
J.~Redmon, S.~Divvala, R.~Girshick, and A.~Farhadi, ``You only look once:
  Unified, real-time object detection,'' in \emph{Proceedings of the IEEE
  conference on computer vision and pattern recognition}, 2016, pp. 779--788.

\bibitem{Jocher2020}
\BIBentryALTinterwordspacing
G.~Jocher, ``{YOLOv5 by Ultralytics},'' 5 2020. [Online]. Available:
  \url{https://github.com/ultralytics/yolov5}
\BIBentrySTDinterwordspacing

\bibitem{Makoviychuk2021}
V.~Makoviychuk, L.~Wawrzyniak, Y.~Guo, M.~Lu, K.~Storey, M.~Macklin,
  D.~Hoeller, N.~Rudin, A.~Allshire, A.~Handa, \emph{et~al.}, ``Isaac gym: High
  performance gpu-based physics simulation for robot learning,'' \emph{arXiv
  preprint arXiv:2108.10470}, 2021.

\bibitem{Kalman1960}
R.~E. Kalman, ``A new approach to linear filtering and prediction problems,''
  1960.

\bibitem{schulman2017}
J.~Schulman, F.~Wolski, P.~Dhariwal, A.~Radford, and O.~Klimov, ``Proximal
  policy optimization algorithms,'' \emph{arXiv preprint arXiv:1707.06347},
  2017.

\end{thebibliography}

\end{document}